\documentclass[11pt,a4paper]{article}
\usepackage{acl2019}
\usepackage{times}
\usepackage{latexsym}
\usepackage{changepage}
\usepackage{graphicx}
\usepackage{boldline}
\usepackage{multirow}
\usepackage{url}
\usepackage{tipa}
\usepackage{chngcntr}
\usepackage{ltablex,booktabs}
\usepackage{subcaption}
\usepackage{capt-of}
\usepackage{lipsum}
\usepackage{float}

\usepackage[bottom]{footmisc}
\aclfinalcopy 

\title{Enhancing Clinical Concept Extraction with Contextual Embeddings}

\author{Yuqi Si \qquad Jingqi Wang \qquad Hua Xu \qquad Kirk Roberts \\School of Biomedical Informatics\\
The University of Texas Health Science Center at Houston
\\
\texttt{\{yuqi.si,kirk.roberts\}}@uth.tmc.edu}

\date{}

\begin{document}
\maketitle
\begin{abstract}

Neural network-based representations (``embeddings") have dramatically advanced natural language processing (NLP) tasks, including clinical NLP tasks such as concept extraction. Recently, however, more advanced embedding methods and representations (\emph{e.g.}, ELMo, BERT) have further pushed the state-of-the-art in NLP, yet there are no common best practices for how to integrate these representations into clinical tasks. The purpose of this study, then, is to explore the space of possible options in utilizing these new models for clinical concept extraction, including comparing these to traditional word embedding methods (word2vec, GloVe, fastText). Both off-the-shelf, open-domain embeddings and pretrained clinical embeddings from MIMIC-III are evaluated. We explore a battery of embedding methods consisting of traditional word embeddings and contextual embeddings, and compare these on four concept extraction corpora: i2b2 2010, i2b2 2012, SemEval 2014, and SemEval 2015. We also analyze the impact of the pre-training time of a large language model like ELMo or BERT on the extraction performance. Last, we present an intuitive way to understand the semantic information encoded by contextual embeddings. Contextual embeddings pre-trained on a large clinical corpus achieves new state-of-the-art performances across all concept extraction tasks. The best-performing model outperforms all state-of-the-art methods with respective F1-measures of 90.25, 93.18 (partial), 80.74, and 81.65. We demonstrate the potential of contextual embeddings through the state-of-the-art performance these methods achieve on clinical concept extraction. Additionally, we demonstrate that contextual embeddings encode valuable semantic information not accounted for in traditional word representations.

\end{abstract}

\section{Introduction}

Concept extraction is the most common clinical natural language processing (NLP) task \cite{tang2013recognizing,kundeti2016clinical,unanue2017recurrent,wang2018clinical}, and a precursor to downstream tasks such as relations \cite{rink2011automatic}, frame parsing \cite{gupta2018automatic, si2018frame}, co-reference \cite{lee2011stanford}, and phenotyping \cite{xu2011extracting, velupillai2018using}. Corpora such as those from i2b2 \cite{uzuner20112010, sun2013evaluating, stubbs2015automated}, ShARe/CLEF \cite{suominen2013overview,kelly2014overview}, and SemEval \cite{pradhan2014semeval,elhadad2015semeval,bethard2016semeval} act as evaluation benchmarks and datasets for training machine learning (ML) models.

Meanwhile, neural network-based representations continue to advance nearly all areas of NLP, from question answering \cite{shen2017word} to named entity recognition \cite{chang2015application,wu2015study,habibi2017deep,unanue2017recurrent,florez2018named} (a close analog of concept extraction). Recent advances in contextualized representations, including ELMo \cite{peters2018deep} and BERT \cite{devlin2018bert}, have pushed performance even further. These have demonstrated that relatively simple downstream models using contextualized embeddings can outperform complex models \cite{seo2016bidirectional} using embeddings such as word2vec \cite{mikolov2013distributed} and GloVe \cite{pennington2014glove}.

In this paper, we aim to explore the potential impact these representations have on clinical concept extraction. Our contributions include the following:
\begin{enumerate}
\item An evaluation exploring numerous embedding methods: word2vec \cite{mikolov2013distributed}, GloVe \cite{pennington2014glove}, fastText \cite{bojanowski2016enriching}, ELMo \cite{peters2018deep}, and BERT \cite{devlin2018bert}.
\item An analysis covering four clinical concept corpora, demonstrating the generalizability of these methods.
\item A performance increase for clinical concept extraction that achieves state-of-the-art results on all four corpora.
\item A demonstration of the effect of pre-training on clinical corpora vs larger open domain corpora, an important trade-off in clinical NLP \cite{roberts2016assessing}.
\item A detailed analysis of the effect of pre-training time when starting from pre-built open domain models, which is important due to the long pre-training time of methods such as ELMo and BERT.
\end{enumerate}

\section{Background}

This section introduces the theoretical knowledge that supports the shift from word-level embeddings to contextual embeddings.

\subsection{Word Embedding Models}
Word-level vector representation methods learn a real-valued vector to represent a single word. One of the most prominent methods for word-level representation is word2vec \cite{mikolov2013distributed}. So far, word2vec has widely established its effectiveness for achieving state-of-the-art performances in a variety of clinical NLP tasks \cite{wang2018comparison}. GloVe \cite{pennington2014glove} is another unsupervised learning approach to obtain a vector representation for a single word. Unlike word2vec, GloVe is a statistical model that aggregates both a global matrix factorization and a local context window. The learning relies on dimensionality reduction on the co-occurrence count matrix based on how frequently a word appears in a context. fastText \cite{bojanowski2016enriching} is also an established library for word representations. Unlike word2vec and GloVe, fastText considers individual words as character n-grams. For instance, \textit{cold} is made of the n-grams \textit{c}, \textit{co}, \textit{col}, \textit{cold}, \textit{o}, \textit{ol}, \textit{old}, \textit{l}, \textit{ld}, and \textit{d}. This approach enables handling of infrequent words that are not present in the training vocabulary, alleviating some out-of-vocabulary issues. 

\begin{figure}[h!]
\centering
\includegraphics[scale=0.32]{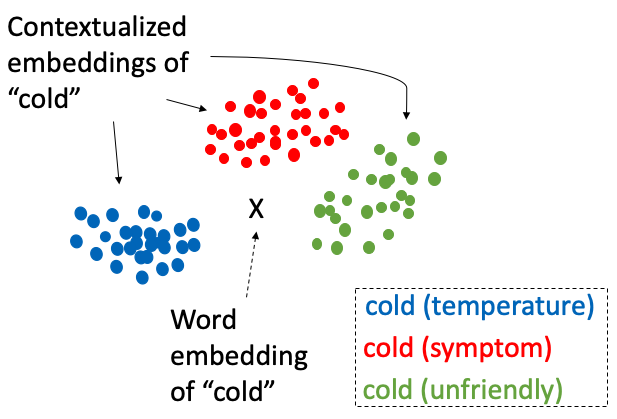}
\caption{Fictional embedding vector points and clusters of ``cold". }
\label{figure:cold_example}
\end{figure} 
However, the effectiveness of word-level representations is hindered by the limitation that they conflate all possible meanings of a word into a single representation and so the embedding is not adjusted to the surrounding context. In order to tackle these deficiencies, advanced approaches have attempted to directly model the word's context into the vector representation.
Figure~\ref{figure:cold_example} illustrates this with the word \textit{cold}, in which a traditional word embedding assigns all senses of the word \textit{cold} with a single vector, whereas a contextual representation varies the vector based on its meaning in context (\emph{e.g.}, cold temperature, medical symptom/condition, an unfriendly disposition). Although a fictional figure is shown here, we later demonstrate this on real data. 

The first contextual word representation that we consider to overcome this issue is ELMo \cite{peters2018deep}. Unlike the previously mentioned traditional word embeddings that constitute a single vector for each word and the vector remains stable in downstream tasks, this contextual word representation can capture the context information and dynamically alter a multilayer representation. At training time, a language model objective is used to learn the context-sensitive embeddings from a large text corpus. The training step of learning these context-sensitive embeddings is known as pre-training. After pre-training, the context-sensitive embedding of each word will be fed into the sentences for downstream tasks. The downstream task learns the shared weights of the inner state of pre-trained language model by optimizing the loss on the downstream task.

BERT \cite{devlin2018bert} is also a contextual word representation model, and, similar to ELMo, pre-training on an 
unlabeled corpus with a language model objective. Compared to ELMo, BERT is deeper in how it handles contextual information due to a deep bidirectional transformer for encoding sentences. It is based on a transformer architecture employing self-attention \cite{vaswani2017attention}. The deep bidirectional transformer is equipped with multi-headed self-attention to prevent locality bias and to achieve long-distance context comprehension. Additionally, in terms of the strategy for how to incorporate these models into the downstream task, ELMo is a feature-based language representation while BERT is a fine-tuning approach. The feature-based strategy is similar to traditional word embedding methods that considers the embedding as input features for the downstream task. The fine-tuning approach, on the other hand, adjusts the entire language model on the downstream task to achieve a task-specific architecture. So while the ELMo embeddings may be used as the input of a downstream model, with the BERT fine-tuning method, the entire BERT model is integrated into the downstream task. This fine-tuning strategy is more likely to make use of the encoded information in the pre-trained language models.

\subsection{Clinical Concept Extraction}
Clinical concept extraction is the task of identifying medical concepts (e.g., problem, test, treatment) from clinical notes. This is typically considered as a sequence tagging problem to be solved with machine learning-based models (e.g., Conditional Random Field) using hand-engineered clinical domain knowledge as features \cite{wang2018clinical, de2011machine}. Recent advances have demonstrated the effectiveness of deep learning-based models with word embeddings as input. Up to now, the most prominent model for clinical concept extraction is a bidirectional Long Short-Term Memory with Conditional Random Field (Bi-LSTM CRF) architecture \cite{habibi2017deep,florez2018named,chalapathy2016bidirectional}. The bidirectional LSTM-based recurrent neural network captures both forward and backward information in the sentence and the CRF layer considers sequential output correlations in the decoding layer using the Viterbi algorithm. 

Most similar to this paper, several recent works have applied contextual embedding methods to concept extraction, both for clinical text and biomedical literature. For instance, ELMo has shown excellent performance on clinical concept extraction \cite{zhu2018clinical}. BioBERT \cite{lee2019biobert} applied BERT primarily to literature concept extraction, pre-training on MEDLINE abstracts and PubMed Central articles, but also applied this model to the i2b2 2010 corpus  without clinical pre-training (we include BioBERT in our experiments
below). A recent preprint by \cite{alsentzer2019publicly} pre-trains on MIMIC-III, similar to our work, but achieves lower performance on the two tasks in common, i2b2 2010 and 2012. Their work does suggest potential value in only pre-training on MIMIC-III discharge summaries, as opposed to all notes, as well as combining clinical pre-training with literature pre-training. Finally, another recent preprint 
proposes the use of BERT not for concept extraction, but for clinical prediction tasks such as 30-day 
readmission prediction \cite{huang2019clinicalbert}.

\section{Methods}
In this paper, we consider both off-the-shelf embeddings from the open domain as well as pretraining clinical domain embeddings on clinical notes from MIMIC-III \cite{johnson2016mimic}, which is a public database of Intensive Care Unit (ICU) patients.

For the traditional word-embedding experiments, the static embeddings are fed into a Bi-LSTM CRF architecture. All words that occur at least five times in the corpus are included and infrequent words are denoted as \texttt{UNK}. To compensate for the loss due of those unknown words, character embeddings for each word are included.

For ELMo, the context-independent embeddings with trainable weights are used to form context-dependent embeddings, which are then fed into the downstream task. Specifically, the context-dependent embedding is obtained through a low-dimensional projection and a highway connection after a stacked layer of a character-based Convolutional Neural Network (char-CNN) and a two-layer Bi-LSTM language model (bi-LM).  Thus, the contextual word embedding is formed with a trainable aggregation of highly-connected bi-LM. Because the context-independent embeddings already consider representation of characters, it is not necessary to learn a character embedding input for the Bi-LSTM in concept extraction. Finally, the contextual word embedding for each word is fed into the prior state-of-the-art sequence labeling architecture, Bi-LSTM CRF, to predict the label for each token.

For BERT, both the BERT\textsubscript{BASE} and BERT\textsubscript{LARGE} off-the-shelf models are used with additional Bi-LSTM layers at the top of the BERT architecture, which we refer to as
BERT\textsubscript{BASE}(\verb|General|) and BERT\textsubscript{LARGE}(\verb|General|), respectively. For background, the BERT authors released two off-the-shelf cased models: BERT\textsubscript{BASE} and BERT\textsubscript{LARGE}, with 110 million and 340 million total parameters, respectively. BERT\textsubscript{BASE} has 12 layers of transformer blocks, 768 hidden units, and 12 self-attention heads, while BERT\textsubscript{LARGE} has 24 layers of transformer blocks, 1024 hidden units, and 16 self-attention heads. So BERT\textsubscript{LARGE} is both ``wider" and ``deeper" in model structure, but is otherwise essentially the same architecture. The models initiated from BERT\textsubscript{BASE}(\verb|General|) and BERT\textsubscript{LARGE}(\verb|General|) are fine-tuned on the downstream task (\emph{e.g.}, clinical concept recognition in our case). Because BERT integrates sufficient label-correlation information, the CRF layer is abandoned and only a Bi-LSTM architecture is used for sequence labeling. Additionally, two clinical domain embedding models are pre-trained on MIMIC-III, initiated from the BERT\textsubscript{BASE} and BERT\textsubscript{LARGE} checkpoints, which we refer to as BERT\textsubscript{BASE}(\verb|MIMIC|) and BERT\textsubscript{LARGE}(\verb|MIMIC|), respectively.

\section{Datasets and Experiments}
\subsection{Datasets}
\begin{table}[h!]
\small
\centering
\renewcommand{\arraystretch}{1.2}
\begin{tabular}{llll}
\hlineB{2.5}
\textbf{Dataset} & \textbf{Subset} & \textbf{\#. Notes} & \textbf{\#. Entities} \\ \hline
\multirow{2}{*}{i2b2 2010} & Train & 349 & 27,837 \\
 & Test & 477 & 45,009 \\\hline
\multirow{2}{*}{i2b2 2012} & Train & 190 & 16,468 \\
 & Test & 120 & 13,594 \\ \hline

 \multirow{2}{*}{\begin{tabular}[c]{@{}l@{}}SemEval 2014 \\ Task 7\end{tabular}} & Train & 199 & 5,816 \\
 & Test & 99 & 5,351 \\\hline
 \multirow{2}{*}{\begin{tabular}[c]{@{}l@{}}SemEval 2015 \\ Task 14\end{tabular}} & Train & 298 & 11,167 \\
 & Test & 133 & 7,998 \\ \hlineB{2.5}
\end{tabular}
\caption{Descriptive statistics for concept extraction datasets}
\label{table:descriptive_dataset}
\end{table}

\begin{table*}[t]
\small
\centering
\setlength{\tabcolsep}{3pt}
\renewcommand{\arraystretch}{1.2}
\begin{tabular}{cccc}
\hlineB{2.5}
\textbf{Method} & \textbf{\begin{tabular}[c]{@{}c@{}}Resource \\ (\#. Tokens/ \#. Vocab)\end{tabular}} & \textbf{Size} & \textbf{\begin{tabular}[c]{@{}c@{}}Language model\end{tabular}} \\ \hline
GloVe & \begin{tabular}[c]{@{}c@{}}Gigaword5 + Wikipedia2014 \\ (6B/ 0.4M)\end{tabular} & 300 & NA \\ \hline
fastText & \begin{tabular}[c]{@{}c@{}}Wikipedia 2017+ UMBC \\webbase corpus  and statmt.org news \\ (16B/ 1M)\end{tabular} & 300 & NA \\ \hline
ELMo & \begin{tabular}[c]{@{}c@{}}WMT 2008-2012 + Wikipedia \\ (5.5B / 0.7M)\end{tabular} & 512 & \begin{tabular}[c]{@{}c@{}}2-layer, 4096-hidden, \\ 93.6M parameters\end{tabular} \\ \hline
BERT\textsubscript{BASE} & \begin{tabular}[c]{@{}c@{}}BooksCorpus+ English Wikipedia \\ (3.3B/ 0.03M\textsuperscript{*})\end{tabular} & 768 & \begin{tabular}[c]{@{}c@{}}12-layer, 768-hidden, 12-heads, \\ 110M parameters\end{tabular} \\ \hline
BERT\textsubscript{LARGE} & \begin{tabular}[c]{@{}c@{}}BooksCorpus + English Wikipedia \\ (3.3B/ 0.03M\textsuperscript{*})\end{tabular} & 1024 & \begin{tabular}[c]{@{}c@{}}24-layer, 1024-hidden, 16-heads, \\ 340M parameters\end{tabular} \\ \hlineB{2.5}
\end{tabular}
\caption{Resources of off-the-shelf embeddings from open domain. (\textsuperscript{*}Vocabulary size calculated after word-piece tokenization)}
\label{table:off-the-shelf}
\end{table*}

Our experiments are performed on four widely-studied shared tasks, the 2010 i2b2/VA challenge \cite{uzuner20112010}, the 2012 i2b2 challenge \cite{sun2013evaluating}, the SemEval 2014  Task 7 \cite{pradhan2014semeval}  and the SemEval 2015 Task 14 \cite{elhadad2015semeval}.  The descriptive statistics for the datasets are shown in Table~\ref{table:descriptive_dataset}.The 2010 i2b2/VA challenge data contains a total of 349 training and 477 testing reports with clinical concept types: {\sc Problem}, {\sc Test} and {\sc Treatment}. The 2012 i2b2 challenge data contains 190 training and 120 testing discharge summaries, with 6 clinical concept types: {\sc Problem}, {\sc Test}, {\sc Treatment}, \hbox{\sc Clinical~department}, {\sc Evidential}, and {\sc Occurrence}. The SemEval 2014 Task 7 data contains 199 training and 99 testing reports with the concept type: \hbox{\sc Disease~disorder}. The SemEval 2015 Task 14 data consists of 298 training and 133 testing reports with the concept type: \hbox{\sc Disease~disorder}.
For the two SemEval tasks, the disjoint concepts are handled with ``BIOHD" tagging schema  \cite{tang2015recognizing}.

The clinical embeddings are trained on MIMIC III \cite{johnson2016mimic}, which consists of almost 2 million clinical notes. Notes that have an \texttt{\small ERROR} tag are first removed, ending up with 1,908,359 notes and 786,414,528 tokens and a vocabulary of size 712,286. For pre-training traditional word embeddings, words are lowercased, as is standard practice. For pre-training ELMo and BERT, casing is preserved.

\subsection{Experimental Setting}

\subsubsection{Concept Extraction}

Concept extraction is based on the model proposed in Lample et al., \shortcite{lample2016neural}, a Bi-LSTM CRF architecture. For traditional embedding methods and ELMo embeddings, we use the same hyperparameters setting: hidden unit dimension at 512, dropout probability at 0.5, learning rate at 0.001, learning rate decay at 0.9, and Adam as the optimization algorithm. Early stopping of training is set to 5 epochs without improvement to prevent overfitting.

\subsubsection{Pre-training of Clinical Embeddings}
Across embedding methods, two different scenarios of pre-training are investigated and compared: 

\begin{enumerate}
\item Off-the-shelf embeddings from the official release, referred to as the \verb|General| model. 
\item Pre-trained embeddings on MIMIC-III, referred to as the \verb|MIMIC| model..
\end{enumerate}

In the first scenario, more details related to the embedding models are shown in in Table~\ref{table:off-the-shelf}. We also apply BioBERT \cite{lee2019biobert}, which is the most recent pre-trained model on biomedical literature initiated from BERT\textsubscript{BASE}. 
\\

In the second scenario, for all the traditional embedding methods, we pre-train 300 dimension embeddings from MIMIC-III clinical notes.  We apply the following hyperparameter settings for all three traditional embedding methods including word2vec, GloVe, and fastText: window size of 15, minimum word count of 5, 15 iterations, and embedding size of 300 to match the off-the-shelf embeddings.

For ELMo, the hyperparameter setting for pre-training follows the default in Peters et al., \shortcite{peters2018deep}. Specifically, a char-CNN embedding layer is applied with 16-dimension character embeddings, filter widths of [1, 2, 3, 4, 5, 6, 7] with respective [32, 32, 64, 128, 256, 512, 1024] number of filters. After that, a two-layer Bi-LSTM with 4,096 hidden units in each layer is added. The output of the final bi-LM language model is projected to 512 dimensions with a highway connection. MIMIC-III was split into a training corpus (80\%) for pre-training and a held-out testing corpus (20\%) for evaluating perplexity. The pre-training step is performed on the training corpus for 15 epochs. The average perplexity on the testing corpus is 9.929.

For BERT, two clinical-domain models initialized from BERT\textsubscript{BASE} and BERT\textsubscript{LARGE} are pre-trained. Unless specified, we follow the authors’ detailed instructions to set up the pre-training parameters, as other options were tested and it has been concluded that this is a useful recipe when pre-training from their released model (\emph{e.g.}, poor model convergence). The vocabulary list consisting of 28,996 word-pieced tokens applied in BERT\textsubscript{BASE} and BERT\textsubscript{LARGE} is adopted.  According to their paper, the performance on the downstream tasks decrease as the training steps increase, thus we decide to save the intermediate checkpoint (every 20,000 steps) and report the performance of intermediate models on the downstream task. 

\begin{table*}[t]
\small
\renewcommand{\arraystretch}{1.3}
\centering
\caption{Test set comparison in exact F1 of embedding methods across tasks. \\ SOTA: state-of-the-art.}
\begin{tabular}{c|cc|cc|cc|cc}
\hlineB{2.5}
\multirow{2}{*}{\textbf{Method}} & \multicolumn{2}{c|}{\textbf{i2b2 2010}} & \multicolumn{2}{c|}{\textbf{i2b2 2012}} & \multicolumn{2}{c|}{\textbf{\begin{tabular}[c]{@{}c@{}}Semeval 2014\\ Task 7\end{tabular}}} & \multicolumn{2}{c}{\textbf{\begin{tabular}[c]{@{}c@{}}Semeval 2015 \\ Task 14\end{tabular}}} \\ \cline{2-9} 
                                 & \texttt{General}    & \texttt{MIMIC}    & \texttt{General}    & \texttt{MIMIC}    & \texttt{General}                              & \texttt{MIMIC}                              & \texttt{General}                                 & \texttt{MIMIC}                                \\ \hline
word2vec                         & 80.38               & 84.32             & 71.07               & 75.09             & 72.2                                          & 77.48                                       & 73.09                                            & 76.42                                         \\
GloVe                            & 84.08               & 85.07             & 74.95               & 75.27             & 70.22                                         & 77.73                                       & 72.13                                            & 76.68                                         \\
fastText                         & 83.46               & 84.19             & 73.24               & 74.83             & 69.87                                         & 76.47                                       & 72.67                                            & 77.85                                         \\
ELMo                             & 83.83               & 87.8              & 76.61               & 80.5              & 72.27                                         & 78.58                                       & 75.15                                            & 80.46                                         \\
BERTbase                         & 84.33               & 89.55             & 76.62               & 80.34             & 76.76                                         & 80.07                                       & 77.57                                            & 80.67                                         \\
BERTlarge                        & 85.48               & \textbf{90.25}    & 78.14               & \textbf{80.91}    & 78.75                                         & \textbf{80.74}                              & 77.97                                            & \textbf{81.65}                                \\
BioBERT                          & 84.76               & -                 & 77.77               & -                 & 77.91                                         & -                                           & 79.97                                            & -                                             \\ \hlineB{1.8}

\multicolumn{1}{c|}{Prior SOTA}  & \multicolumn{2}{c|}{88.60 \cite{zhu2018clinical}}            & \multicolumn{2}{c|}{\textbf{\textdoublebarpipe} \cite{liu2017entity}}                   & \multicolumn{2}{c|}{80.3 \cite{tang2015recognizing}}                                                                    & \multicolumn{2}{c}{81.3 \cite{zhang2014uth_ccb}}                                                                      \\ \hline

\hlineB{2.5}
\end{tabular}

\scriptsize{ \textbf{\textdoublebarpipe} The SOTA on the i2b2 2012 task is only reported in partial-matching F1. That result, 92.29 \cite{liu2017entity}, is below the equivalent we achieve on {partial-matching} F1 with BERT\textsubscript{LARGE}(\texttt{MIMIC}), 93.18.}

\label{table:result comparison}
\end{table*}

\subsubsection{Fine-tuing BERT}
Fine-tuning the BERT clinical models on the downstream task requires some adjustments. First, instead of randomly initializing the Bi-LSTM output weights, Xavier initialization is utilized \cite{glorot2010understanding}, without which the BERT fine-tuning failed to converge (this was not necessary for ELMo). Second, early stopping of fine-tuning is set to 800 steps without improvement to prevent overfitting. Finally, post-processing steps are conducted to align the BERT output with the concept gold standard, including handling truncated sentences and word-pieced tokenization.

\subsubsection{Evaluation and Computational Costs}
10\% of the official training set is used as a development set and the official test set is used to report performance. The specific performance metrics are precision, recall, and F1-measure for exact matching. The pre-training BERT experiments are implemented in TensorFlow \cite{abadi2016tensorflow} on a NVIDIA Tesla V100 GPU (32G), other experiments are performed on a NVIDIA Quadro M5000 (8G). The time for pre-training ELMo, BERT\textsubscript{BASE} and BERT\textsubscript{LARGE} for every 20K checkpoint is 4.83 hours, 3.25 hours, and 5.16 hours, respectively. These models were run until manually set to stop at 320K iterations (82.66 hours $\sim$ roughly 3.4 days), 700K iterations (113.75 hours $\sim$ roughly 4.7 days), and 700K iterations (180.83 hours $\sim$ roughly 7.5 days), respectively.

\section{Results}
\subsection{Performance Comparison}
The performance on the respective test sets for the embedding methods on the four clinical concept extraction tasks are reported in Table~\ref{table:result comparison}. The performance is evaluated in exact matching F1. In general, embeddings pre-trained on the clinical corpus performed better than the same method pre-trained on an open domain corpus.  

For i2b2 2010, the best performance is achieved by BERT\textsubscript{LARGE}(\verb|MIMIC|) with an F1 of 90.25. It improves the performance by 5.18 over the best performance of the traditional embeddings achieved by GloVe (\verb|MIMIC|) with an F1 of 85.07. As expected, both ELMo and BERT clinical embeddings outperform the off-the-shelf embeddings with relative increase up to 10\%. 

The best performance on the i2b2 2012 task is achieved by BERT\textsubscript{LARGE}(\verb|MIMIC|) with an F1 of 80.91 across all the alternative methods. It increases F1 by 5.64 over GloVe(\verb|MIMIC|), which obtains the best score (75.27) among the traditional embedding methods. As expected, ELMo and BERT with pre-trained clinical embeddings exceed the off-the-shelf open domain models.

The most efficient model for SemEval 2014 task achieved an exact matching with F1 of 80.74 by BERT\textsubscript{LARGE}(\verb|MIMIC|). Notably, traditional embedding models pre-trained on clinical corpus such as GloVe(\verb|MIMIC|) obtained a higher performance than contextual embedding model trained on open domain, namely ELMo(\verb|General|).

For the SemEval 2015 task, as the experiments are performed only in concept extraction, the models are evaluated using the official evaluation script from the SemEval 2014 task. Note the training set (298 notes) for the SemEval 2015 task is the training (199 notes) and test set (99 notes) combined for the SemEval 2014 task. The best performance on the 2015 task is achieved by BERT\textsubscript{LARGE}(\verb|MIMIC|) with an F1 of 81.65.

The detailed performance for each entity category including {\sc Problem}, {\sc Test} and \hbox{\sc Treatment} on the 2010 task is shown in Table~\ref{table:10mimic}. Both ELMo and BERT show improvements to all three categories, with ELMo outperforming the traditional embeddings on all three, and BERT outperforming ELMo on all three. One notable aspect with BERT is that {\sc Treatment}s see a larger jump: {\sc Treatment} is the lowest-performing category for ELMo and the traditional embeddings despite there being slightly more {\sc Treatment}s than {\sc Test}s in the data, but for BERT the {\sc Treatment}s category outperforms {\sc Test}s.

\begin{table}[h!]
\centering
\renewcommand{\arraystretch}{1.5}
\setlength{\tabcolsep}{1.7pt}
\scriptsize
\begin{tabular}{c|c c c c cc}
\hlineB{2.5}
 & \textbf{word2vec} & \textbf{GloVe} & \textbf{fastText} & \textbf{ELMo} & \textbf{BERT\textsubscript{BASE}} & \textbf{BERT\textsubscript{LARGE}} \\ \hline
{\sc Problem} & 84.16 & 85.08 & 84.32 & 88.76 & \textbf{89.61} & 89.26 \\
{\sc Test} & 85.93 & 84.96 & 84.01 & 87.39 & 88.09 & \textbf{88.8} \\
{\sc Treatment} & 83.14 & 84.73 & 83.89 & 86.98 & 88.3 & \textbf{89.14} \\ \hlineB{2.5}
\end{tabular}
\caption{Performance of each label category with \hbox{pre-trained} MIMIC models on i2b2 2010 task.}
\label{table:10mimic}
\end{table}

Table~\ref{table:12mimic} shows the results for each event type on the 2012 task with embeddings pre-trained from MIMIC- III. Generally, the biggest improvement by the contextual embeddings over the traditional embeddings is achieved on the {\sc Problem} type (BERT\textsubscript{LARGE}: 86.1, GloVe: 77.83). This is reasonable because in clinical notes, diseases and conditions normally appear in certain types of surrounding context with similar grammar structures. Thus, it is necessarily important to take advantage of contextual representations to capture the surrounding context for that particular concept. Interestingly, ELMo outperforms both BERT models for \hbox{\sc Clinical~department} and {\sc Occurrence}.

\begin{table}[h!]
\scriptsize
\centering
\renewcommand{\arraystretch}{1.2}
\setlength{\tabcolsep}{1.7pt}
\begin{tabular}{c|cccccc}
\hlineB{2.5}
& \textbf{word2vec}& \textbf{GloVe} & \textbf{FastText} & \textbf{ELMo} & \textbf{BERT\textsubscript{BASE}} & \textbf{BERT\textsubscript{LARGE}} \\ \hline
{\sc Problem} & 76.49 & 77.83 & 75.35 & 84.1 & 85.91 & \textbf{86.1}  \\
{\sc Test} & 78.12 & 81.26 & 76.94 & 84.76 & \textbf{86.88} & 86.56 \\
{\sc Treatment} & 76.22 & 78.52 & 76.88 & 83.9 & 84.27 & \textbf{85.09}  \\
\hbox{\sc Clinical~dept} & 78.18 & 77.92 & 77.27 & \textbf{83.71} & 77.92 & 78.23 \\
{\sc Evidential} & 73.14 &  74.26 & 72.94 & 72.95 & 74.21 & \textbf{74.96}\\
{\sc Occurrence} & 64.77 & 64.19 & 61.02 & \textbf{66.27} & 62.36 & 65.65 \\ \hlineB{2.5}
\end{tabular}
\caption{Performance of each label category with pre-trained MIMIC models on i2b2 2012 task.}
\label{table:12mimic}

\end{table}
\subsection{Pre-training Evaluation}
The efficiency of pre-trained ELMo and BERT models are investigated by reporting the loss during pre-training steps and by evaluating the intermediate checkpoints on downstream tasks. It is observed for both ELMo and BERT at their pre-training stages, the train perplexity or loss decreases as the steps increase, indicating that the language model is actually adapting to the clinical corpus. If there is no intervention to stop the pre-training process, it will lead to a very small loss value. However, this will ultimately cause overfitting on the pre-training corpus (shown in Supplemental Figure 1). However, this will bring to another common issue that the model might be overfitting on the training set. 

\begin{figure}[h!]
\centering
\includegraphics[scale=0.6]{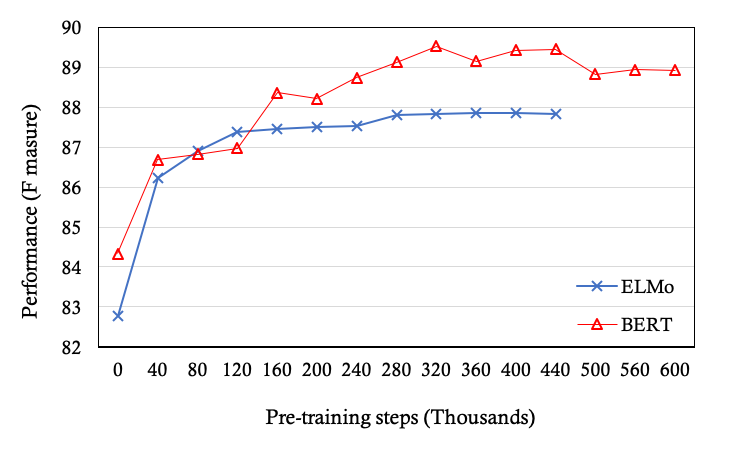}
\caption{Performances on the i2b2 2010 task governed by the steps of pre-training epochs on ELMo (\texttt{MIMIC}) and BERT\textsubscript{BASE} (\texttt{MIMIC}) .}
\label{figure:pretraining}
\end{figure}

Using i2b2 2010 as the downstream task, the final performance at each intermediate checkpoint of the pre-trained model is shown in Figure~\ref{figure:pretraining}. For ELMo, as the pre-training proceeds, the performance of the downstream task remains stable after a certain number of iterations (the maximum F1 reaches 87.80 at step 280K). For BERT\textsubscript{BASE}, the performance on the downstream task is less steady and tends to decrease after achieving its optimal model, with the maximum F1 89.55 at step 340K. We theorize that this is due to initializing the MIMIC model with the open-domain BERT model: over many iterations on the MIMIC data, the information learned from the large open corpus (3.3 billion words) is lost and would eventually converge on a model similar to one initialized from scratch. Thus, limiting pre-training on a clinical corpus to a certain number of iterations provides a useful trade-off, balancing the benefits of a large open-domain corpus while still learning much from a clinical corpus. We hope this is a practical piece of guidance for the clinical NLP community when they intend to generate their own pre-trained model from a clinical corpus.

\section{Discussion}

This study explores the effects of numerous embedding methods on four clinical concept extraction tasks. Unsurprisingly, domain-specific embedding models outperform open-domain embedding models. All types of embeddings enable consistent gains in concept extraction tasks when pre-trained on a clinical domain corpus. Further, the contextual embeddings outperform traditional embeddings in performance. Specifically, large improvements can be achieved by pre-training a deep language model from a large corpus, followed by a task-specific fine-tuning. 

\subsection{State-of-the-art Comparison}
Among the four clinical concept extraction corpora, the i2b2 2012 task reports the partial matching F1 as the organizers reported in Sun et al., \shortcite{sun2013evaluating} and the other three tasks report the exact matching F1. Currently, the state-of-the-art models in i2b2 2010, i2b2 2012, SemEval 2014 task 7, and Semeval 2015 Task 14 are reported with F1 of 88.60 \cite{zhu2018clinical}, 92.29 \cite{liu2017entity}, 80.3 \cite{tang2015recognizing} and 81.3 \cite{zhang2014uth_ccb}, respectively. With the most advanced language model representation method pretrained on a large clinical corpus, namely BERT \textsubscript{LARGE}(\verb|MIMIC|), we achieved new state-of-the-art performances across all tasks. \hbox{BERT \textsubscript{LARGE}(\verb|MIMIC|)} outperform the state-of-the-art models on all four tasks with respective F-measures of 90.25, 93.18 (\textit{partial F1}), 80.74, and 81.65. 
\subsection{Semantic Information from Contextual Embeddings}
Here, we explore the semantic information captured by the contextual representation and infer that the contextual embedding can encode information that a single word vector fails to. First, we select 30 sentences from both web texts and clinical notes in which the word \textit{cold} appears (The actual sentences can be found in Supplemental Table 1). The embedding vectors of \textit{cold} in 30 sentences from four embedding models, ELMo(\verb|General|), ELMo(\verb|MIMIC|), \hbox{BERT\textsubscript{LARGE}(\verb|General|)}, and \hbox{BERT \textsubscript{LARGE}(\verb|MIMIC|)}, were derived. This results in 120 vectors for the same word across four embeddings. For each embedding method, Principal Component Analysis (PCA) is performed to reduce the dimensionality to 2.

\begin{figure*}[t!]
\centering
\includegraphics[scale=0.42]{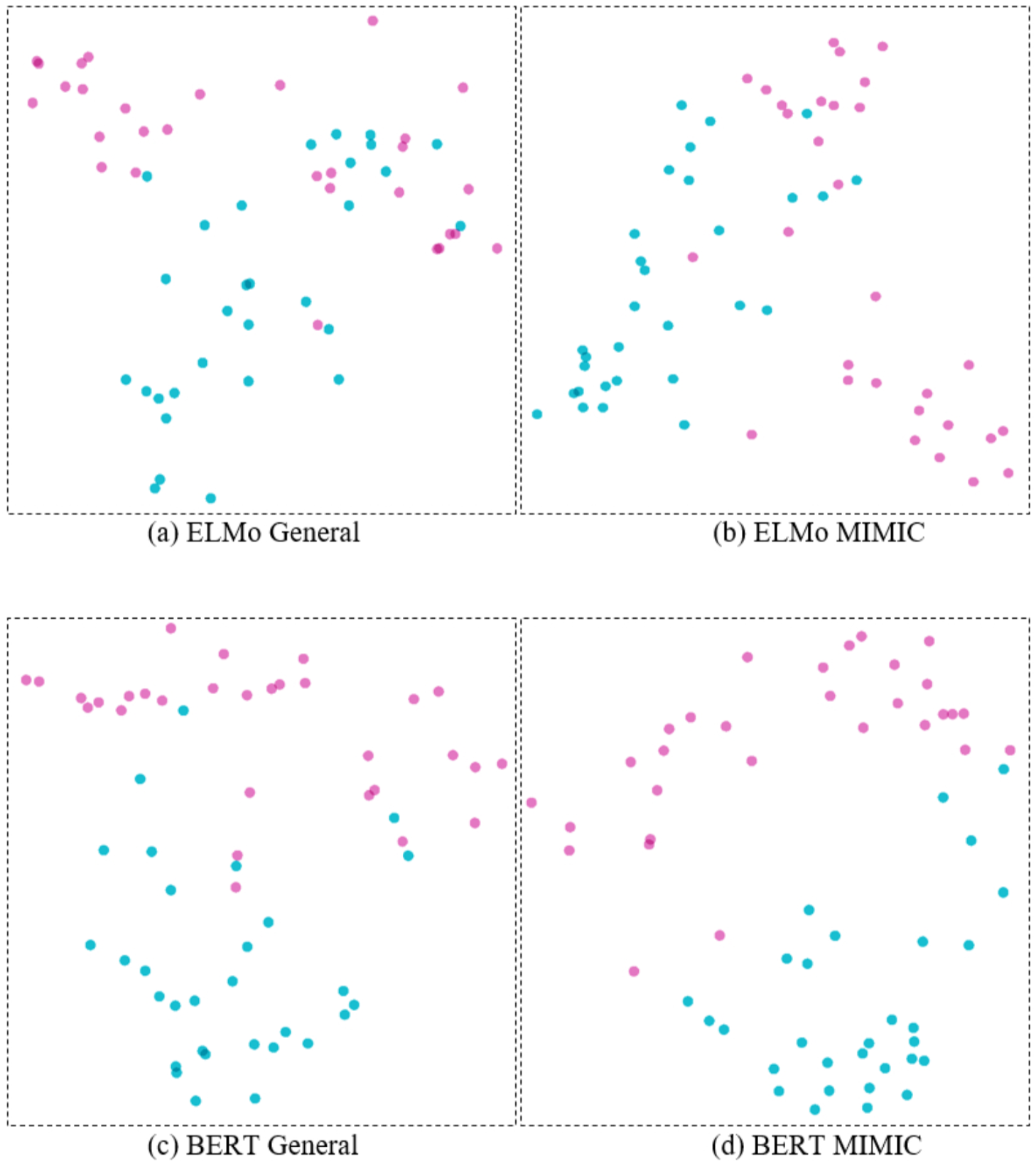}
\caption{Principal Component Analysis (PCA) visualizations using embedding vectors of cold from embedding models (purple: \textit{cold} as temperature meaning; red: \textit{cold} as symptom).}
\label{figure:cold_pca}
\end{figure*}

The PCA visualizations are shown in Figure~\ref{figure:cold_pca}. As expected, the vectors of \textit{cold} generated by \hbox{ELMo(\texttt{General})} are mixed within two different meaning labels. The vectors generated by \hbox{BERT\textsubscript{LARGE}(\verb|General|)} and \hbox{BERT \textsubscript{LARGE}(\verb|MIMIC|)} are more clearly clustered into two groups. \hbox{ELMo(\texttt{General})} is unable to discriminate the different meanings of the word \textit{cold}, specifically between temperature and symptom. The visualization result is also consistent with the performance on the concept extract tasks where \hbox{ELMo(\texttt{General})} tends to get a poorer performance compared with the other three models.

Second, traditional word embeddings are commonly evaluated using lexical similarity tasks, such as those by \cite{pakhomov2010semantic,pakhomov2016corpus}, which compare two words outside any sentence-level context. While not entirely appropriate for comparing contextual embeddings such as ELMo and BERT because the centroid of the embedding clusters are not necessarily meaningful, such lexical similarity tasks do provide motivation for investigating the clustering effects of lexically similar (and dissimilar) words. In Supplemental Figure 2, we compare four words from \cite{pakhomov2016corpus}: \textit{tylenol}, \textit{motrin}, \textit{pain}, and \textit{herpes} based on 50-sentence samples from MIMIC-III and the same 2-dimension PCA visualization technique. One would expect \textit{tylenol} (acetaminophen) and \textit{motrin} (ibuprofen) to be similar, and in fact the clusters overlap almost completely. Meanwhile, \textit{pain} is a nearby cluster, while \textit{herpes} is quite distant. So while contextual embeddings are not well-suited to context-free lexical similarity tasks, the aggregate effects (clusters) still demonstrate similar spatial relationships as traditional word embeddings.

\subsection{Lexical Segmentation in BERT}

One important notable difference between BERT and both ELMo and the traditional word embeddings is that BERT breaks words down into sub-word tokens, referred to as wordpieces \cite{schuster2012japanese}. This is accomplished via statistical analysis on a large corpus, as opposed to using a morphological lexicon. The concern for clinical NLP, then, is if a different word piece tokenization method is appropriate for clinical text as opposed to general text (i.e., books and Wikipedia for the pre-trained BERT models). Supplemental Table 2 shows the word piece tokenization for the medical words from the lexical similarity corpus developed by \cite{pakhomov2016corpus}. The results do not exactly conform to traditional medical term morphology (e.g., ``appendicitis" is broken into ``app", ``-end", ``-icit", ``-is", as opposed to having the suffix ``-itis"). Note that this isn't necessary a bad segmentation: it is possible this would outperform a word piece tokenization based on the SPECIALIST lexicon \cite{browne2000specialist}. What is not in dispute, however,is that further experimentation is required, such as determining word pieces from MIMIC-III. Note this is not as simple as it at first seems. The primary issue is that the BERT models we use in this paper were first pre-trained on a 3.3 billion word open-domain corpus, then fine-tuned on MIMIC-III. Performing word piece tokenization on MIMIC-III would at a minimum require repeating the pre-training process on the open-domain corpus (with the clinical word pieces) in order to get comparable embedding models. Given the range of experimentation necessary to determine the best word piece strategy, we leave this experimentation to future work.

\section{Conclusion}
In this paper, we present an analysis of different word embedding methods and investigate their effectiveness on four clinical concept extraction tasks. We compare between traditional word representation methods as well as the advanced contextual representation methods. We also compare pre-trained contextual embeddings using a large clinical corpus against the performance of off-the-shelf pre-trained models on open domain data. Primarily, the efficacy of contextual embeddings over traditional word vector representations are highlighted by comparing the performances on clinical concept extraction. Contextual embeddings also provide interesting semantic information that is not accounted for in traditional word representations. Further, our results highlight the benefits of embeddings through unsupervised pre-training on clinical text corpora, which achieve higher performance than off-the-shelf embedding models and result in new state-of-the-art performance across all tasks.

\section*{Acknowledgments}

This work was supported by the U.S. National Institutes of Health (NIH) and the Cancer Prevention and Research Institute of Texas (CPRIT). Specifically, NIH support comes from the National Library of Medicine (NLM) under award R00LM012104 and R01LM010681, as well as the National Cancer Institute under award U24CA194215. CPRIT support for computational resources was provided under awards RP170668 and RR180012.

\bibliography{context_embed}
\bibliographystyle{acl_natbib}

\end{document}